\PassOptionsToPackage{unicode}{hyperref}
\PassOptionsToPackage{hyphens}{url}
\PassOptionsToPackage{dvipsnames,svgnames,x11names}{xcolor}
\documentclass[
  10pt,
]{article}
\usepackage{xcolor}
\usepackage[margin=0.85in]{geometry}
\usepackage{amsmath,amssymb}
\usepackage{iftex}
\ifPDFTeX
  \usepackage[T1]{fontenc}
  \usepackage[utf8]{inputenc}
  \usepackage{textcomp} 
\else 
  \usepackage{unicode-math} 
  \defaultfontfeatures{Scale=MatchLowercase}
  \defaultfontfeatures[\rmfamily]{Ligatures=TeX,Scale=1}
\fi
\usepackage{lmodern}
\ifPDFTeX\else
\fi
\IfFileExists{upquote.sty}{\usepackage{upquote}}{}
\IfFileExists{microtype.sty}{
  \usepackage[]{microtype}
  \UseMicrotypeSet[protrusion]{basicmath} 
}{}
\makeatletter
\@ifundefined{KOMAClassName}{
  \IfFileExists{parskip.sty}{%
    \usepackage{parskip}
  }{
    \setlength{\parindent}{0pt}
    \setlength{\parskip}{6pt plus 2pt minus 1pt}}
}{
  \KOMAoptions{parskip=half}}
\makeatother
\usepackage{longtable,booktabs,array}
\usepackage{calc} 
\usepackage{etoolbox}
\makeatletter
\patchcmd\longtable{\par}{\if@noskipsec\mbox{}\fi\par}{}{}
\makeatother
\IfFileExists{footnotehyper.sty}{\usepackage{footnotehyper}}{\usepackage{footnote}}
\makesavenoteenv{longtable}
\usepackage{graphicx}
\makeatletter
\newsavebox\pandoc@box
\newcommand*\pandocbounded[1]{
  \sbox\pandoc@box{#1}%
  \Gscale@div\@tempa{\textheight}{\dimexpr\ht\pandoc@box+\dp\pandoc@box\relax}%
  \Gscale@div\@tempb{\linewidth}{\wd\pandoc@box}%
  \ifdim\@tempb\p@<\@tempa\p@\let\@tempa\@tempb\fi
  \ifdim\@tempa\p@<\p@\scalebox{\@tempa}{\usebox\pandoc@box}%
  \else\usebox{\pandoc@box}%
  \fi%
}
\def\fps@figure{htbp}
\makeatother
\ifLuaTeX
\usepackage[bidi=basic]{babel}
\else
\usepackage[bidi=default]{babel}
\fi
\babelprovide[main,import]{american}

\def\languageshorthands#1{}
\ifLuaTeX
  \usepackage[english]{selnolig} 
\fi
\usepackage[]{natbib}
\usepackage{booktabs}
\usepackage{bookmark}
\IfFileExists{xurl.sty}{\usepackage{xurl}}{} 
\hypersetup{
  pdftitle={How Weight Encoding Affects Language Model Placement and Performance on the Apple Neural Engine},
  pdfauthor={Shahir M A},
  pdflang={en-US},
  colorlinks=true,
  linkcolor={Maroon},
  filecolor={Maroon},
  citecolor={Blue},
  urlcolor={Blue},
  pdfcreator={LaTeX via pandoc}}

\title{How Weight Encoding Affects Language Model Placement and
Performance on the Apple Neural Engine}
\author{Shahir M A}
\date{24 September 2026}

\begin{document}
\maketitle
\begin{abstract}
Weight compression can alter accelerator placement as well as memory
traffic, complicating the interpretation of inference speedups. We
investigate this interaction on the Apple Neural Engine through the
public Core ML deployment path. Five independently trained
language-model checkpoints span two architectures and dense fp16, int8,
and ternary weights encoded with two-bit lookup tables. We combine
compiler device plans, synchronized memory-controller measurements, and
compute-unit exclusion controls for a fixed single-token forward
workload. On an M1, the smaller fp16 export executes on the CPU despite
permitting ANE execution, whereas its compressed counterparts exhibit
ANE activity. The int8 export reduces warm forward latency by a factor
of 1.9. The larger fp16 export also uses the ANE, indicating that dense
encoding alone does not determine placement. Separate M3 energy
measurements support the same direction of change. These results
establish encoding-dependent placement in the measured deployment stack
and show that compression comparisons require joint measurement of
backend selection, latency, and traffic.
\end{abstract}

Independent researcher ·
\href{mailto:shahir@thesmartlanguage.com}{\nolinkurl{shahir@thesmartlanguage.com}}

\section{Introduction}\label{introduction}

Weight compression is commonly evaluated through model size and
inference latency, but these measurements can conflate two effects:
reduced data movement on a given accelerator and a change in the backend
executing the model. This distinction matters on Apple's public Core ML
path, where a request permitting the Apple Neural Engine (ANE) also
permits CPU fallback. Successful conversion and prediction therefore do
not establish that an exported language model executes on the ANE.

The deployment question is whether compressed and dense exports of a
given architecture use the same backend, and how their observed latency
relates to memory traffic. Compiler device plans provide evidence about
intended placement, but they do not directly measure execution.
Conversely, system-wide hardware counters record activity without
identifying the model responsible. Resolving the relationship requires
compiler information and runtime measurements supported by
device-exclusion and idle controls.

We study five trained exports from two convolution-dominant
language-model families, comparing dense fp16, int8, and ternary weights
stored in two-bit lookup tables. On the measured M1 stack, the smaller
fp16 export produces no ANE byte traffic and has similar latency with
and without ANE permission. Its compressed counterparts produce ANE
traffic and complete the same workload faster; excluding the ANE removes
that signal. The larger fp16 export also produces ANE traffic. The
observed placement difference therefore depends on the deployed model
and encoding, rather than a general exclusion of dense weights.

The study combines designed graph probes, trained-model comparisons, and
synchronized hardware telemetry. We quantify traffic against an explicit
mixed-width weight inventory and use a matched normalization control to
separate intermediate dtype from arithmetic formulation. Within each
family, the encoding arms share architecture and training settings but
contain independently trained weights. The results characterize these
deployed artifacts under a fixed single-token forward workload; they do
not isolate re-encoding of identical learned values. Separate M3 energy
measurements provide evidence for the placement effect on a second chip.

\section{Related work and scope}\label{related-work-and-scope}

Bryngelson's monograph studies ANE architecture, compiler behavior,
compression, and language-model execution \citep{bryngelson}. ANEForge,
Orion, and the mixture-of-experts study explore direct programming and
execution on Apple neural hardware \citep{aneforge, orion, moe}. Our
narrower focus is the public Core ML deployment path, treating encoding
as an experimental variable and comparing planned devices with
controlled telemetry. Different compiler paths and generations limit
mechanistic comparisons between these studies.

Hardware-aware architecture selection already includes the ANE.
MobileNetV4 evaluates mobile architectures across heterogeneous devices
and introduces Mobile MQA \citep{mobilenet}; EdgeDiT targets on-device
diffusion transformers \citep{edgedit}. Designed microbenchmarks also
have a methodological precedent in nn-Meter's detection of runtime
execution kernels \citep{nnmeter}. Talaria provides an interactive
workflow for model cost and optimization decisions \citep{talaria}.

Apple's transformer deployment articles describe layout and operator
choices for the ANE \citep{apple2022, apple2024}. Its Core ML
optimization guide recommends palettization for ANE deployment
\citep{appleopt}. ANEMLL is an existing conversion and deployment
pipeline and is credited by our legacy normalization builder
\citep{anemll}. The community neural-engine reference discusses
determining whether a model uses the ANE, and more-ane-transformers
reports language-model implementations and timing results
\citep{hollance, moreane}. Our telemetry contribution is the controlled
use of specific byte channels, including group/unit validation and
negative controls.

The workloads use familiar components: RMSNorm, gated feed-forward
layers, rotary position embeddings, and grouped-query attention
\citep{rmsnorm, swiglu, rope, gqa}. The low-bit training arm follows the
ternary-weight motivation of BitNet b1.58 and uses straight-through
optimization \citep{bitnet, ste}. TinyStories supplies the training task
\citep{tinystories}. We use trained weights and check numerical
agreement before measuring deployment performance.

\section{Methodology}\label{methodology}

The experimental design combines operation-level eligibility probes with
model-level placement and traffic measurements. An initial survey
compiled designed graphs containing convolutions, normalization,
attention, and embedding lookup, recording supported and preferred
devices. We then trained and exported language models under multiple
weight encodings and measured each compiled artifact with the ANE
permitted or excluded. The principal analysis comprises five trained
exports, synchronized counter windows, and a matched
intermediate-precision control. The original graph survey and
operator-mix timings remain in the reproduction artifact.

\begin{longtable}[]{@{}
  >{\raggedright\arraybackslash}p{(\linewidth - 6\tabcolsep) * \real{0.2500}}
  >{\raggedright\arraybackslash}p{(\linewidth - 6\tabcolsep) * \real{0.2500}}
  >{\raggedright\arraybackslash}p{(\linewidth - 6\tabcolsep) * \real{0.2500}}
  >{\raggedright\arraybackslash}p{(\linewidth - 6\tabcolsep) * \real{0.2500}}@{}}
\caption{Experimental comparisons. Encoding arms use separately trained
weights; device controls reuse each compiled model.
\label{tab:instruments}}\tabularnewline
\toprule\noalign{}
\begin{minipage}[b]{\linewidth}\raggedright
Comparison
\end{minipage} & \begin{minipage}[b]{\linewidth}\raggedright
Held fixed
\end{minipage} & \begin{minipage}[b]{\linewidth}\raggedright
Changed
\end{minipage} & \begin{minipage}[b]{\linewidth}\raggedright
Recorded
\end{minipage} \\
\midrule\noalign{}
\endfirsthead
\toprule\noalign{}
\begin{minipage}[b]{\linewidth}\raggedright
Comparison
\end{minipage} & \begin{minipage}[b]{\linewidth}\raggedright
Held fixed
\end{minipage} & \begin{minipage}[b]{\linewidth}\raggedright
Changed
\end{minipage} & \begin{minipage}[b]{\linewidth}\raggedright
Recorded
\end{minipage} \\
\midrule\noalign{}
\endhead
\bottomrule\noalign{}
\endlastfoot
Weight encoding & Architecture within each family; input & Training
precision and exported weights & Device plan, latency, bytes \\
Device exclusion & Compiled model and input & CPU only, CPU+ANE, CPU+GPU
& Latency and counter signal \\
Normalization dtype & Arithmetic; fp16 input/output & fp16 or fp32
intermediates & Eligibility and output error \\
\end{longtable}

\subsection{Model construction and
training}\label{model-construction-and-training}

We constructed small models from scratch using the LFM2 implementation
in Transformers \citep{lfm2}. Each architecture combines short
convolution layers with a final grouped-query attention layer. The q25
family has 9 convolution layers, hidden width 512, feed-forward width
1536, and 8 query heads; q50 has 11 convolution layers, hidden width
640, feed-forward width 1920, and 10 query heads. Both use 2 key/value
heads, convolution width 3, and a tied input embedding and output
projection. The family names refer approximately to non-embedding
parameter counts; the table reports total parameters.

\begin{longtable}[]{@{}lllll@{}}
\caption{Principal checkpoints and mixed-width inventory. A zero
quantized count denotes the dense fp16 arm.
\label{tab:models}}\tabularnewline
\toprule\noalign{}
Model & Hidden / layers & Total M & Quantized M & Accounted MB \\
\midrule\noalign{}
\endfirsthead
\toprule\noalign{}
Model & Hidden / layers & Total M & Quantized M & Accounted MB \\
\midrule\noalign{}
\endhead
\bottomrule\noalign{}
\endlastfoot
q25-fp16 & 512 / 10 & 27.943 & 0.000 & 55.886 \\
q25-int8 & 512 / 10 & 27.943 & 25.821 & 30.065 \\
q25-ternary & 512 / 10 & 27.943 & 25.821 & 10.699 \\
q50-fp16 & 640 / 12 & 51.155 & 0.000 & 102.311 \\
q50-ternary & 640 / 12 & 51.155 & 48.497 & 17.442 \\
\end{longtable}

We trained each model on 100,000 stories from the TinyStories V2 GPT-4
training split and used 5,000 stories from its validation split
\citep{tinystories}. Training covered 3 epochs, totaling 61.05 million
tokens processed per model. Training configurations and validation
metrics are retained in the artifact.

\subsection{Core ML export and compiler
plans}\label{core-ml-export-and-compiler-plans}

We re-exported fp16, int8, and ternary checkpoints from q25, and fp16
and ternary checkpoints from q50. We loaded each checkpoint in
evaluation mode, used eager attention with a traceable causal mask, and
disabled the persistent key/value cache. The exporter generated a fixed
token input with shape \([1,1]\), saved its Torch output, and traced the
forward pass. Core ML conversion used an int32 token input and fp16
compute.

Before conversion, the compressed checkpoints' linear weights were
rounded onto their exact discrete grids, and the simulated
activation-quantization wrappers were removed. We then left dense
weights in fp16, applied symmetric int8 weight encoding to the int8 arm,
or used unique-value palettization for the ternary arm. The ternary
representation uses two-bit indices into a palette of scaled weight
values. Rounding before palettization prevents floating-point training
residuals from creating extra distinct entries. The experiment therefore
tests exported weight encodings with fp16 compute, rather than
end-to-end integer arithmetic.

We saved each model package, compiled it for Core ML, and recorded the
full compute plan. We classify an operation as ANE-eligible when its
supported-device set includes the ANE, and as ANE-preferred when the
compiler selects that device. Our preferred-operation counts include
only assigned operations and exclude unassigned constant-reconstruction
nodes. These counts describe the plan, not a fraction of runtime or
arithmetic work.

The M1 measurements were collected on 10 September 2026 with macOS 26.1
(25B78), Python 3.13.2, coremltools 9.0, PyTorch 2.10.0, and
Transformers 4.57.6. The export logs retain compatibility warnings and
hashes of checkpoints, packages, and input fixtures. These fresh exports
are distinct from the earlier packages measured in Appendix A.

\subsection{Inference workload and compute-unit
controls}\label{inference-workload-and-compute-unit-controls}

For each compiled model, we launched a separate worker for CPU\_ONLY,
CPU\_AND\_NE, and CPU\_AND\_GPU. The same compiled model and fixed input
were used across its three workers. CPU\_AND\_NE permits the Neural
Engine but also permits CPU fallback; CPU\_AND\_GPU excludes the Neural
Engine and provides a second negative control. We reversed the device
order on alternating models and ran workers sequentially.

Each worker loaded the model and performed twenty warmup calls. Before
timing, we compared its output logits with the saved Torch reference,
requiring finite values and cosine similarity of at least 0.999. The
smallest observed cosine was 0.99997361. We then measured ten windows of
one thousand completed predictions for CPU\_AND\_NE, and three windows
of two hundred fifty predictions for each exclusion setting. Every call
was timed, and the final output of each window was checked for finite
values.

The workload repeatedly evaluates a fixed, seeded token input without a
persistent key/value cache or autoregressive feedback. Reported latency
therefore characterizes a warm, sequence-one forward pass.
Autoregressive generation throughput requires a separate evaluation with
state updates and token selection.

\subsection{Counter acquisition and
aggregation}\label{counter-acquisition-and-aggregation}

We measured ANE memory traffic (read + write bytes, abbreviated RD+WR)
through macOS IOReport. On the M1, our sampler subscribed to the AMC
Stats group and identified the ANE RD and ANE WR byte counters, which
record memory reads and writes associated with ANE activity. We also
captured ANE DCS RD and ANE DCS WR for comparison. All four channels had
to match the expected group, name, and byte unit; missing, duplicate, or
negative readings invalidated the measurement.

A separate sampler process took counter snapshot \(A\) after model
warmup and acknowledged it before the worker started a measurement
window. The worker then completed \(N\) predictions and signaled the
sampler to take snapshot \(B\). We computed traffic per prediction as

\[
\text{bytes per prediction}
=\frac{(\mathrm{RD}_B-\mathrm{RD}_A)+(\mathrm{WR}_B-\mathrm{WR}_A)}{N}.
\]

Here \(N\) is the actual number of completed calls in the window. We
summed only the ANE RD and ANE WR deltas; the DCS channels were retained
separately because their observed deltas mirrored this accounting. These
counters measure system-wide traffic associated with ANE activity,
including possible weight, activation, and runtime transfers. They do
not identify individual weight accesses.

Before and after each worker, we required three consecutive one-second
idle samples with zero RD+WR traffic, allowing at most fifteen attempts
per gate. An initial complete capture had residual idle and CPU-only
traffic. We retained it under \texttt{headline} and repeated the
measurements under \texttt{headline\_isolated}, reusing the compiled
models, removing a Core ML import from the supervisor, and applying the
consecutive-zero gates. The original residual's source was not
independently established; the two captures are not pooled.

Bandwidth uses the sampler's elapsed interval, including the handshake
overhead. Latency uses the per-call clock: each reported value is the
median of the window medians. Window ranges describe variation within
this session. The compute-unit nulls test whether the byte signal
depends on allowing the ANE; the idle sequences record whether unrelated
activity was present around each worker.

\subsection{Matched intermediate-precision
control}\label{matched-intermediate-precision-control}

To isolate the precision difference identified in the graph survey, we
constructed four programs: RMSNorm and addition, each with fp16 or fp32
intermediates. Every program used fp16 input/output of shape
\([1,2048]\). For RMSNorm we held the arithmetic sequence, unit scale,
and epsilon fixed, using the same fp16-representable approximation to
\(10^{-4}\) in both arms. We disabled conversion passes that could
silently lower fp32 intermediates. Addition provided a simple control
for the same precision change.

We evaluated zero input and three random seeds at scales 0.1, 1, and 10
under CPU\_ONLY and CPU\_AND\_NE. Outputs were checked against a float64
reference, requiring finite values, maximum absolute error at most 0.02,
and relative L2 error at most 0.01; zero input used the absolute
criterion alone. We recorded operand dtypes and supported/preferred
devices for each operation. The resulting comparison measures the effect
of intermediate precision on compiler eligibility while checking
numerical agreement.

\newpage

\section{Encoding-dependent
placement}\label{encoding-dependent-placement}

\begin{longtable}[]{@{}lllll@{}}
\caption{M1 measurements. Entries are medians of per-window median warm
prediction times; requested units permit fallback.
\label{tab:latency}}\tabularnewline
\toprule\noalign{}
Model & ANE-pref. / assigned & CPU ms & CPU+ANE ms & CPU+GPU ms \\
\midrule\noalign{}
\endfirsthead
\toprule\noalign{}
Model & ANE-pref. / assigned & CPU ms & CPU+ANE ms & CPU+GPU ms \\
\midrule\noalign{}
\endhead
\bottomrule\noalign{}
\endlastfoot
q25-fp16 & 0/331 & 1.454 & 1.458 & 3.256 \\
q25-int8 & 324/335 & 1.419 & 0.770 & 8.631 \\
q25-ternary & 324/335 & 1.508 & 0.638 & 16.255 \\
q50-fp16 & 380/387 & 2.787 & 2.079 & 5.071 \\
q50-ternary & 380/391 & 2.757 & 0.909 & 19.667 \\
\end{longtable}

The q25 fp16 export has 0/331 assigned operations preferring the ANE and
produces zero measured traffic over 10000 CPU\_AND\_NE predictions. Its
CPU\_ONLY and CPU\_AND\_NE latencies are close. The int8 and ternary
exports have 324/335 and 324/335 ANE-preferred operations, respectively,
produce substantial traffic, and complete the same fixed-input workload
faster when the ANE is permitted. For each quantized model, excluding
the ANE removes the byte signal. Together, the plan, latency contrast,
and compute-unit nulls support execution on different backends for these
deployed artifacts.

\begin{figure}
\centering
\includegraphics[width=1\linewidth,height=\textheight,keepaspectratio]{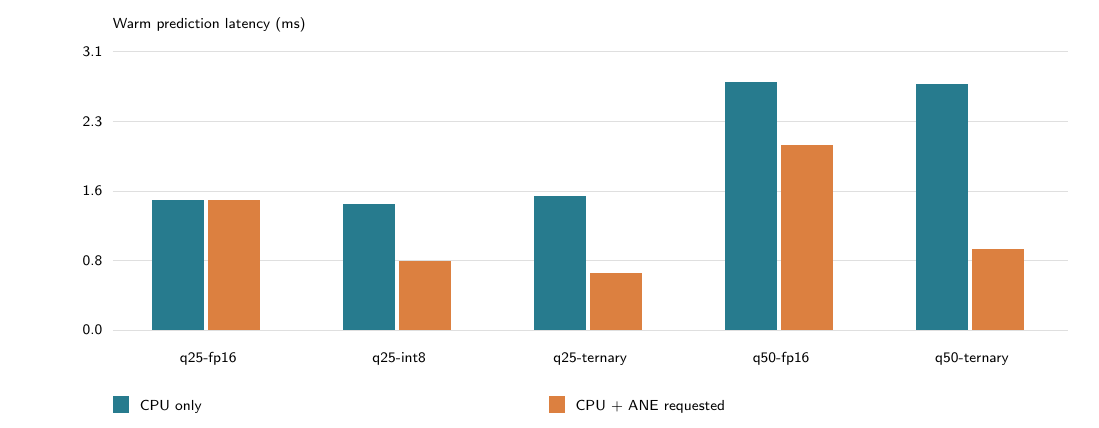}
\caption{Matched-architecture deployment comparison. Bars are
window-median latency summaries, not generation
throughput.}\label{fig:placement}
\end{figure}

The larger fp16 model also produces ANE traffic, so fp16 weights alone
do not explain the placement difference. A compiler cost model trading
compute, transfer, and launch costs is a plausible explanation. Deriving
a crossover would require validated graph partition costs and transfer
accounting for the tested compiler.

Across 80 windows and 57500 measured predictions, all 30 CPU/GPU
exclusion windows read zero, as do the final three samples of every
accepted idle gate. Of 118 idle samples collected while waiting for
those gates, 28 were nonzero and are retained. The protocol therefore
establishes quiet intervals around each worker, including after loading
and teardown activity.

\newpage

\subsection{M3 measurements}\label{m3-measurements}

\begin{longtable}[]{@{}lll@{}}
\caption{M3 placement and compute-unit controls. Energy Model/ANE
readings in millijoules indicate accelerator activity; they are not
energy per prediction. \label{tab:m3}}\tabularnewline
\toprule\noalign{}
M3 cell & ANE energy mJ & Prediction ms \\
\midrule\noalign{}
\endfirsthead
\toprule\noalign{}
M3 cell & ANE energy mJ & Prediction ms \\
\midrule\noalign{}
\endhead
\bottomrule\noalign{}
\endlastfoot
q25-fp16 & 0 & 1.125 \\
q25-int8 & 2200 & 0.645 \\
int8 / CPU+ANE & 2148 & 0.640 \\
int8 / CPU only & 0 & 1.130 \\
int8 / CPU+GPU & 0 & 3.760 \\
\end{longtable}

We measured the q25 fp16 and int8 exports on an 8 GB M3 MacBook Air
running macOS 26.5.2 and coremltools 9.0 on 30 July 2026, with one run
per cell. We read the Energy Model/ANE channel in millijoules because
the M1 AMC byte channels are absent. The retained measurements do not
include synchronized prediction counts and sampling durations, so the
energy readings establish accelerator activity but do not support an
energy-per-prediction comparison. The fp16 model has no reported ANE
energy, while int8 has a nonzero signal and lower latency. The
compute-unit control gives zero ANE energy when the ANE is excluded.
This smaller summary capture supports the encoding-dependent behavior on
a second chip, using energy rather than the M1 byte-window protocol. It
does not replicate the byte ratios. The different telemetry requires
checking channel group and unit on each chip.

\section{Traffic and mixed-width
accounting}\label{traffic-and-mixed-width-accounting}

Let \(P\) be total checkpoint parameters and \(Q\) the number assigned
to the quantized weight arm. For index or scalar width \(b\), define the
accounted weight inventory

\[W_{\mathrm{accounted}}=Qb/8+2(P-Q).\]

This includes fp16 weights left outside quantization. It excludes
palette metadata, packing overhead, intermediate activations, and
unknown runtime allocations. Consequently we call it \textbf{accounted
weight bytes}, rather than an exact deployed footprint. For each
synchronized window with traffic \(D\) and completed prediction count
\(N\), we report \(D/N\) and compare it with this explicit inventory. MB
and GB use decimal units.

\begin{longtable}[]{@{}lllll@{}}
\caption{M1 ANE memory traffic (read + write bytes) for CPU\_AND\_NE.
Ratios use the accounted weight bytes in the checkpoint table; the zero
cell is a placement null. \label{tab:bytes}}\tabularnewline
\toprule\noalign{}
Model & MB / call & Window range MB & Ratio to inventory & GB/s \\
\midrule\noalign{}
\endfirsthead
\toprule\noalign{}
Model & MB / call & Window range MB & Ratio to inventory & GB/s \\
\midrule\noalign{}
\endhead
\bottomrule\noalign{}
\endlastfoot
q25-fp16 & 0.000 & 0.000--0.000 & 0.000 & 0.00 \\
q25-int8 & 28.463 & 28.461--28.476 & 0.947 & 36.25 \\
q25-ternary & 11.134 & 11.132--11.144 & 1.041 & 17.31 \\
q50-fp16 & 102.795 & 102.790--102.799 & 1.005 & 48.00 \\
q50-ternary & 17.931 & 17.927--17.936 & 1.028 & 19.62 \\
\end{longtable}

\begin{figure}
\centering
\includegraphics[width=1\linewidth,height=\textheight,keepaspectratio]{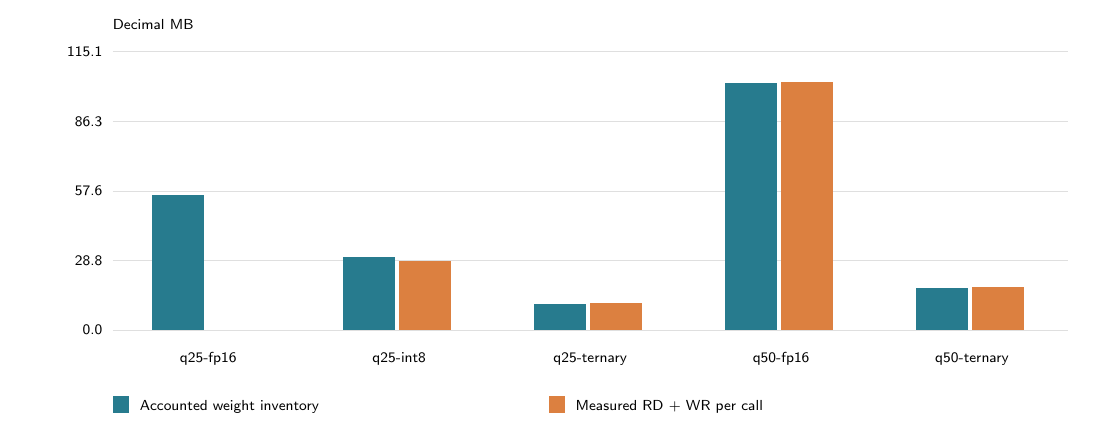}
\caption{Accounted weight inventory and observed traffic. The q25 fp16
cell has no byte signal; the other ratios are descriptive accounting
comparisons.}\label{fig:bytes}
\end{figure}

For the four traffic-positive cells, the median ratio spans
0.947--1.041. The q25 fp16 placement null is excluded from this
comparison. Aggregate traffic includes possible activation and runtime
overhead; proximity to the weight inventory does not establish that each
weight is read once.

The int8 q25 export streams 1.019 bytes per total parameter, compared
with 2.009 for the traffic-positive fp16 q50 export. This is compatible
with a bandwidth benefit from compressed deployment, although the models
differ in size. Bryngelson describes an M1 path that expands int8 to
dense fp16 before streaming \citep[chap.~7]{bryngelson}. Our
observations motivate a comparison between deployment paths; neither
aggregate traffic nor exported \texttt{constexpr\_*} reconstruction
nodes locate the backend's expansion step.

The two-bit graphs are admitted and produce ANE traffic. Their fp16
embedding already raises bytes per parameter above a pure two-bit width.
The measurements consequently establish admission without identifying
physical index width or arithmetic precision.

\section{Eligibility: a matched dtype
control}\label{eligibility-a-matched-dtype-control}

The matched programs isolate intermediate precision while holding the
normalization arithmetic fixed. The table reports compiler eligibility
and the maximum errors across the inputs and requested devices described
in the methodology.

\begin{longtable}[]{@{}
  >{\raggedright\arraybackslash}p{(\linewidth - 8\tabcolsep) * \real{0.2000}}
  >{\raggedright\arraybackslash}p{(\linewidth - 8\tabcolsep) * \real{0.2000}}
  >{\raggedright\arraybackslash}p{(\linewidth - 8\tabcolsep) * \real{0.2000}}
  >{\raggedright\arraybackslash}p{(\linewidth - 8\tabcolsep) * \real{0.2000}}
  >{\raggedright\arraybackslash}p{(\linewidth - 8\tabcolsep) * \real{0.2000}}@{}}
\caption{Matched dtype controls. Error maxima are against the numerical
reference across the retained inputs and requested devices.
\label{tab:rms}}\tabularnewline
\toprule\noalign{}
\begin{minipage}[b]{\linewidth}\raggedright
Graph
\end{minipage} & \begin{minipage}[b]{\linewidth}\raggedright
ANE eligible / assigned
\end{minipage} & \begin{minipage}[b]{\linewidth}\raggedright
Max abs. error
\end{minipage} & \begin{minipage}[b]{\linewidth}\raggedright
Max rel. L2
\end{minipage} & \begin{minipage}[b]{\linewidth}\raggedright
Checks
\end{minipage} \\
\midrule\noalign{}
\endfirsthead
\toprule\noalign{}
\begin{minipage}[b]{\linewidth}\raggedright
Graph
\end{minipage} & \begin{minipage}[b]{\linewidth}\raggedright
ANE eligible / assigned
\end{minipage} & \begin{minipage}[b]{\linewidth}\raggedright
Max abs. error
\end{minipage} & \begin{minipage}[b]{\linewidth}\raggedright
Max rel. L2
\end{minipage} & \begin{minipage}[b]{\linewidth}\raggedright
Checks
\end{minipage} \\
\midrule\noalign{}
\endhead
\bottomrule\noalign{}
\endlastfoot
rmsnorm fp16 & 7/7 & 0.002362 & 0.000530 & 20 \\
rmsnorm fp32 & 0/8 & 0.000975 & 0.000283 & 20 \\
add fp16 & 2/2 & 0.000000 & 0.000000 & 20 \\
add fp32 & 0/3 & 0.000000 & 0.000000 & 20 \\
\end{longtable}

All 80 numerical checks pass. Intermediate fp32 excludes the tested
operations from ANE eligibility, while fp16 is eligible. The tiny graphs
can still prefer the CPU. This supports a dtype explanation for the
original contrast and agrees with Bryngelson's documented fp16
activation constraint \citep[table 17.1]{bryngelson}.

The probe records operand dtypes and supported/preferred-device
responses. The tested public SDK and Python DeviceUsage wrapper expose
no reason field; unavailable reasons are recorded as null. The retained
SDK header documents that interface.

The earlier shape matrix and its separate numerical checks remain in the
artifact as descriptive evidence for the recorded shapes and
environment.

\section{Operator choice and a practical design
procedure}\label{operator-choice-and-a-practical-design-procedure}

Earlier operator-mix measurements, retained in Appendix A, motivate
comparing architectures at similar deployment sizes. The nearby ternary
timings suggest candidates for further evaluation, rather than a
statistical tie. Grouped-query attention reduces projection parameters
for the chosen head counts; placement and latency must still be measured
for the resulting exports.

A practical procedure follows from the measurements. First fix the
actual workload, including shape, cache behavior, and quality
requirement. Enumerate and validate available telemetry channels and
their units. Export candidate architectures under multiple encodings;
check output agreement before measuring speed. Read eligibility and
preferred devices separately, then bracket counted predictions with idle
and compute-unit controls. Compare latency and byte traffic with an
explicit inventory that includes unquantized weights. Finally, evaluate
language quality and end-to-end generation separately before choosing a
deployment.

\section{Corrections to version 1}\label{corrections-to-version-1}

Version 1 attributed a normalization eligibility difference to
formulation and reported a constant 0.77 fraction of nominal weight
width. We withdraw those interpretations: the normalization comparison
was confounded by dtype, and the byte calculation used stale values and
incomplete mixed-width accounting. The synchronized measurements here
supersede the earlier byte values, whose aggregate logs lack sufficient
per-channel provenance for a unique retrospective correction.
Preferred-operation shares also now exclude unassigned reconstruction
nodes. Detailed changes and artifact corrections are recorded in the
repository's revision notes.

\section{Limitations and
reproducibility}\label{limitations-and-reproducibility}

The inference experiment covers one M1 session, five separately trained
checkpoints, and a fixed sequence-one input. It establishes export
agreement on that input, not equal task quality across models or
performance on long contexts, stateful generation, cold starts, and
sustained thermal workloads. The M3 evidence is a smaller energy-based
capture. Results are specific to the recorded devices and software
stacks.

Compute plans describe compiler choices, while system-wide counters
provide aggregate activity. The exclusion controls and idle gates
strengthen attribution but cannot rule out another process active only
during a measurement window. The counters do not identify physical
storage formats or individual weight accesses.

The reproduction artifact under
\texttt{results/2026-09-10\_paper\_v2\_release/} contains raw
measurements, numerical checks, configurations, hashes, and the sources
needed to regenerate every table, figure, and the paper. Rebuilding
requires no model weights or ANE hardware. Inference reruns require the
separately retained checkpoints and the recorded Apple software stack.
Commands, dependencies, and the distinction between rebuilding results,
rerunning inference, and repeating training are documented in
\texttt{paper/v2/README.md} and the bundle README.

\section{Conclusion}\label{conclusion}

In the measured public Core ML deployment path, changing weight encoding
accompanies a change from a CPU-executed model to ANE activity,
supported by compiler preferences, latency, byte counters and exclusion
controls. Explicit dtype controls and mixed-width accounting narrow two
earlier interpretations. The useful result is a reproducible connection
between exported representation and observed execution, with its limits
visible in the measurements.

\newpage

\section*{Appendix A. Earlier export
timings}\label{appendix-a.-earlier-export-timings}
\addcontentsline{toc}{section}{Appendix A. Earlier export timings}

The table preserves all retained sequence-one export measurements. These
are best-of-loop timings and are not pooled with the repeated-window
medians in the main experiment. The a-series changes
attention/convolution mix; the q-series is the convolution-dominant
baseline. Checkpoint configuration manifests accompany the rows. These
timings do not establish equal task quality or a statistical tie between
architectures.

\begin{longtable}[]{@{}llll@{}}
\caption{Earlier sequence-one export timings. Package size differs from
the accounted weight inventory; replication is weaker than in the main
experiment. \label{tab:historical}}\tabularnewline
\toprule\noalign{}
Export & Package MB & Best CPU+ANE ms & Best CPU ms \\
\midrule\noalign{}
\endfirsthead
\toprule\noalign{}
Export & Package MB & Best CPU+ANE ms & Best CPU ms \\
\midrule\noalign{}
\endhead
\bottomrule\noalign{}
\endlastfoot
q50-fp16 & 102.5 & 1.903 & 2.543 \\
q50-int8 & 51.5 & 1.105 & 2.502 \\
q25-fp16 & 56.0 & 1.346 & 1.349 \\
q50-ternary & 17.6 & 0.865 & 2.524 \\
q25-int8 & 28.2 & 0.765 & 1.350 \\
a25-dense-fp16 & 49.0 & 1.103 & 1.119 \\
a25-dense-ternary & 10.0 & 0.639 & 1.116 \\
a25-alt-ternary & 10.5 & 0.629 & 1.238 \\
a05-alt-ternary & 3.5 & 0.331 & 0.360 \\
a50-alt-ternary & 16.8 & 0.859 & 2.243 \\
a50-alt-fp16 & 96.0 & 1.902 & 2.330 \\
a05-alt-fp16 & 12.0 & 0.342 & 0.345 \\
\end{longtable}

\renewcommand\refname{References}
\bibliography{references.bib}

\end{document}